\documentclass{article}
\usepackage{spconf,amsmath,graphicx,amsfonts}
\usepackage{algorithm}
\usepackage[noend]{algpseudocode}

\usepackage{hhline}
\usepackage{multirow}
\usepackage{makecell}

\usepackage{booktabs}

\usepackage{enumitem}
\setlist{nosep, leftmargin=14pt}

\usepackage{mwe} 

\title{DISYRE: Diffusion-Inspired SYnthetic REstoration for Unsupervised Anomaly Detection}
\name{\parbox{0.9\linewidth}{\centering Sergio Naval Marimont$^{1}$ \qquad Matthew Baugh$^{2}$\qquad Vasilis Siomos$^{1}$ \qquad Christos Tzelepis$^{1}$ \quad Bernhard Kainz$^{2,3}$
\qquad Giacomo Tarroni$^{1,2}$}}

\address{$^{1}$ CitAI Research Centre, Department of Computer Science City, University of London \\
$^{2}$ BioMedIA, Department of Computing, Imperial College London \\
$^{3}$ Department of Artificial Intelligence in Biomedical Engineering, FAU Erlangen-Nürnberg}
\begin{document}
%
\maketitle
\begin{abstract}

Unsupervised Anomaly Detection (UAD) techniques aim to identify and localize anomalies without relying on annotations, only leveraging a model trained on a dataset known to be free of anomalies.
Diffusion models learn to modify inputs $x$ to increase the probability of it belonging to a desired distribution, \emph{i.e.}, they model the score function $\nabla_x \log p(x)$.
Such a score function is potentially relevant for UAD, since $\nabla_x \log p(x)$ is itself a pixel-wise anomaly score.
However, diffusion models are trained to invert a corruption process based on Gaussian noise and the learned score function is unlikely to generalize to medical anomalies. 
This work addresses the problem of how to learn a score function relevant for UAD and proposes DISYRE: Diffusion-Inspired SYnthetic REstoration.
We retain the diffusion-like pipeline but replace the Gaussian noise corruption with a gradual, synthetic anomaly corruption so the learned score function generalizes to medical, naturally occurring anomalies.
We evaluate DISYRE on three common Brain MRI UAD benchmarks and substantially outperform other methods in two out of the three tasks.

\end{abstract}

\section{Introduction} 
\label{sections:introduction}

\begin{figure*}[ht!]
    \centering
    \includegraphics[width=\textwidth]{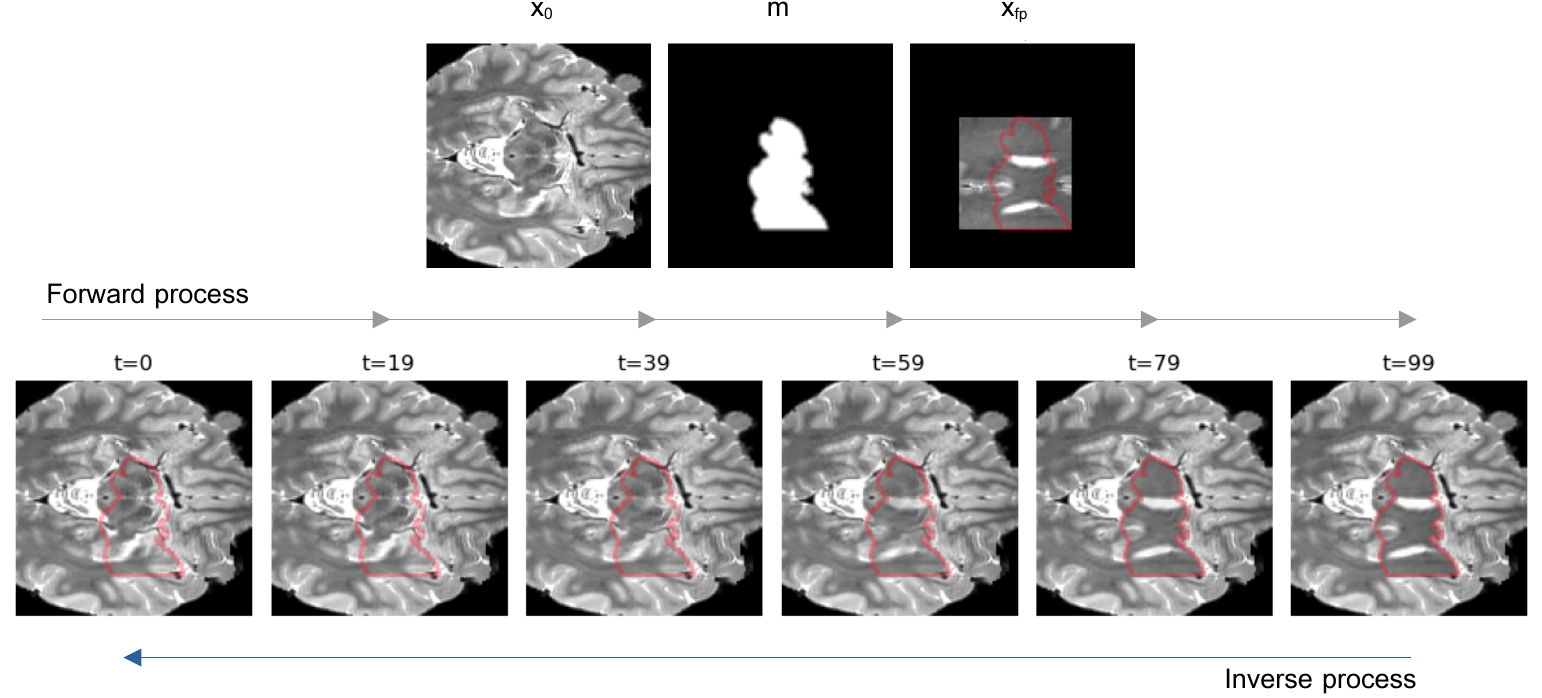}
    \caption{1st row: examples of anomaly generation inputs: healthy image $x_0$, shape mask $m$ and foreign patch $x_{fp}$. 2nd row: synthetic corruption forward process for different $t$ values (with $T=100$).}
    \label{fig:processes}
\end{figure*}


Unsupervised Anomaly Detection (UAD) in the field of medical image analysis involves the detection and/or localization of irregularities in medical images without the need for annotations.
This approach is particularly valuable since obtaining annotations can be both expensive and challenging~\cite{Litjens2017}. 
During the prediction phase, UAD methodologies employ comparisons between a test image and the established distribution, often using a parameterized model of this distribution.
Anomalies are detected when deviations from this healthy distribution are observed in the test image.


Many UAD methods rely on modelling the expected distribution with generative models such as variational auto-encoders \cite{Baur2020} or generative adversarial networks \cite{Schlegl2019}.
These rely on the assumption that generative models trained on in-distribution data will be unable to reconstruct out-of-distribution regions, meaning that the reconstruction error can be used to identify anomalous pixels.
Unfortunately, this assumption often fails, as models may not accurately reconstruct normal data and the remaining residual error needs to be calibrated as subject-specific bias~\cite{Baur2020,phanes}.
Furthermore, auto-encoders are also known to have the ability to also reconstruct anomalies \cite{phanes}.


As an alternative, restoration-based methods~\cite{Chen2020} use Maximum-A-Posteriori (MAP) estimation to remove local anomalies from images.
Many approaches have since followed this idea, often using a two-stage restoration process; first identifying anomalous areas or latent variables and then resampling them to become in-distribution~\cite{Naval2020,phanes}.


More recently, diffusion models have emerged as the state-of-the-art method for image generation. 
The most common of these is the Denoising Diffusion Probabilistic Model (DDPM)~\cite{DDPM}, which constructs a mapping from Gaussian noise to a desired image distribution.
To do this, DDPMs leverage Langevin dynamics~\cite{DDPM} to learn a score function, i.e. $\nabla_x \log p(x)$, which is able to iteratively denoise a sample of Gaussian noise to produce a sample from the target image distribution.
Such a score function is potentially very relevant for UAD, as it is able to indicate both where and how to edit the image to increase its likelihood, directly acting as a pixel-wise anomaly score.
However, as a standard DDPM learns a score function that is specific to images corrupted with Gaussian noise, it does not generalize to correcting images containing other types of anomalies.

Because of this, DDPM's score function has not been favoured when applying diffusion models to UAD.
AnoDDPM~\cite{anoddpm} instead replaces Gaussian noise with simplex noise, as the multi-scale nature of simplex noise trains the score function to be able to correct larger anomalous structures.
This is consistent with what is reported by DAE~\cite{DAE} authors, who found that denoising models trained on reasonably coarse noise outperformed those using pixel-level noise. An alternative approach 
\cite{WollebJ20222} involves utilizing classifier guidance to encourage the model to restore the image to be part of an expected distribution. However, this requires training a discriminator model with sample-level labels.


Separately from generative models, Foreign Patch Interpolation (FPI)~\cite{fpi} proposed directly training anomaly segmentation networks by adding synthetic anomalies to otherwise healthy anatomies.
The anomalies are generated by interpolating a square region of the sample with a patch extracted from a separate sample, controlling the interpolation with a factor $\alpha$.
One of the limitations of FPI is that it produces image gradients on the boundaries between synthetic anomalies and original images, a feature which the model could exploit as a shortcut to identify synthetic anomalies.
Poisson Image Interpolation (PII)~\cite{PII} proposes to use Poisson image blending to mitigate this by performing the interpolation on the gradients of the image, seamlessly blending the anomaly with the surrounding area.
Naval et al.~\cite{MOOD22} took a simpler approach, softening the interpolation factor at the boundaries to make a more gradual change, as well as generating anomalies with randomly generated shapes to produce greater variety in the generated anomalies.

\noindent\textbf{Contribution:} In this work we introduce DISYRE: Diffusion-Inspired SYnthetic REstoration. By replacing the Gaussian noise corruption in diffusion models with a synthetic anomaly degradation process allows for directly using a diffusion-like model for unsupervised anomaly detection and localisation. It does this by learning a score function that generalizes to naturally-occurring medical anomalies and indicates how to modify a test image to bring it within the healthy distribution. Diffusion-like pipelines without noise have been previously proposed for image generation \cite{cold_diffusion}, but to our knowledge this is the first approach specifically designed for UAD. 

\section{Method} 
\label{sections:method}

\noindent\textbf{Synthetic anomaly generation:}
We adopt the synthetic anomaly generation mechanism from~\cite{MOOD22}, where a section of the image is corrupted replacing the original pixel intensities with a linear interpolation of original intensities and a foreign patch extracted randomly from the training set. Local synthetic anomalies are introduced by
\begin{equation}
x_t = (1 - \alpha \cdot m) \cdot x_0 + \alpha \cdot m \cdot x_{fp}
\label{eq:xt}
\end{equation}
where $x_0 \in \mathbb{R}^{H \times W}$ denotes the original (healthy) image, $x_{fp} \in \mathbb{R}^{H \times W}$ denotes the foreign patch extracted from a healthy sample different from $x_0$, $\alpha \in [0,1]$ denotes the interpolation factor that controls the convex combination of $x_0$ and $x_{fp}$ within the synthetic anomaly, and $m \in \mathbb{R}^{H \times W}$ denotes the mask component that both controls the shape of the synthetic anomaly and softens the interpolation factor towards the edges of the anomalous region. Note that $m_i \in [0,1]$.

\noindent\textbf{Forward and backward corruption process:}
In our diffusion-inspired proposed framework, we define a schedule to gradually corrupt the images increasing the $\alpha$ factor defined in Eq.~\ref{eq:xt} and train a model to revert the modified forward corruption process. We note that we do not use random noise, although synthetic anomalies are randomly generated.

Following the DDPM literature, we define the schedule $\alpha_t$ to increase with time steps $t$, \emph{i.e.} when $t=0$, $\alpha_t = 0$ and the image belongs to the expected distribution, while when $t=T$, $\alpha_t = 1$ and the image is highly anomalous. We use the DDPM \cite{DDPM} schedule, however in our case it defines the interpolation of the original image and foreign patch instead of original image and noise. Note that our $\alpha_t$ is analogous to $1-\sqrt{\overline{\alpha_t}}$ in the diffusion literature. Given that we want interpolations $x_t$ to remain in the range $[0,1]$, we use a convex combination of image and foreign patch, so the weighting of the original image becomes $(1-\alpha_t)$. In Fig.~\ref{fig:processes}, we include an example of the synthetic corruption process that shows similarities with the standard diffusion process where textures / high frequencies are corrupted at small $t$ values and structures / low frequencies are corrupted at high $t$ values.

We train our model $P_\theta$ to restore corrupted images $x_t$ into their healthy counterparts when conditioning on $t$, $x_0 \approx \hat{x}_0 = P_{\theta}(x_t,t)$.
Network parameters are therefore optimized using the objective $\mathbb{E}_{t \sim [1,T],x_0,x_{fp}} (\| x_0 - P_{\theta}(x_t,t) \|^2)$.

\noindent\textbf{Anomaly Score:} At inference time, we propose to use the trained model $P_{\theta}$ to \textit{restore} test images. This process is similar to the standard sampling of a DDPM~\cite{DDPM}, but crucially different in that we do not start from noise, but instead from a test image. Additionally, instead of iterating through all $T$ steps,  we skip steps based on a $step\_size$ hyper-parameter.

In order to localize anomalies we propose an Anomaly Score ($AS$) defined as the accumulation of the absolute gradients across all the restoration steps, so it is sensitive to the different degrees of abnormalities present in the corruption process. 
Algorithm~\ref{alg:restoration} describes our \textit{healing} procedure, with $P$ as the trained model $P_{\theta}$ and $Q$ the process specified in Eq. \ref{eq:xt} which corrupts $\hat{x}_o$, interpolating it with the \textit{unhealed} $x_t$ using $t_\text{next}$ and $m=1$.
In addition to the $AS$, the algorithm produces a restoration $\hat{x}_0$, a \textit{healed} version of $x_t$.
In our experiments we also evaluate using directly single-step restorations residuals, \emph{i.e.}, $\lvert P(x_t,t) - x_0 \rvert$, as anomaly score conditioning on different $t$ values. 

\begin{algorithm}
\caption{Inference: Image restoration}\label{alg:restoration}
\begin{algorithmic}[1]
\State $x_t \gets x$
\State $t \gets T$
\State $AS \gets \text{zeros\_like}(x)$
\While {true}
    \State $\hat{x}_0 \gets P(x_t, t)$
    \State $AS \gets AS + |\hat{x}_0 - x_t|$
    \If{$t_{next} \leq 0$}
        \State \textbf{break} \Comment{Exit the loop}
    \EndIf
    \State $t_{next} \gets t - step\_size$
    \State $x_t \gets Q(\hat{x}_0,x_t,t_{next}, m=1)$
    \State $t \gets t_{next}$
\EndWhile
\State \textbf{end}
\end{algorithmic}
\end{algorithm}

During training we use a patch-based pipeline. At inference time we use sliding window inference to obtain predictions for full axial slices. We hypothesized that the model is more accurate on the patches with a bigger proportion of foreground vs background. Consequently, we propose to weight patch predictions by the foreground percentage prior to combine patch-predictions and report results both without (DISYRE) and with foreground weighting (DISYRE f.w.).
\section{Experiments} 
\label{sections:results}

\noindent\textbf{Experimental setup: }
We adopt the experimental setup for Brain MRI from~\cite{UPD}, which includes the following datasets:
\begin{itemize}
    \item Cambridge Centre for Ageing and Neuroscience dataset (CamCAN) \cite{CamCAN}: T1- and T2-weighted images of 652 healthy, lesion-free, adult subjects. 
    \item Anatomical Tracings of Lesions After Stroke (ATLAS) dataset \cite{ATLAS}: T1-weighted scans of stroke patients ($N=655$). This dataset contains annotated lesions after stroke.  
    \item Multimodal Brain Tumor Segmentation (BraTS) Challenge dataset \cite{brats} 2020 edition: We used the T1- and T2-weighted images from the challenge training set. The dataset provides manual segmentations of gliomas.

\end{itemize}

We use the CamCAN dataset for training ($N=602$), holding out a set with $N=50$ subjects to identify training convergence.
BraTS and ATLAS are used to evaluate UAD performance.
We follow the registration and normalization protocols provided in~\cite{UPD}, training our model using patches of $128 \times 128$ pixels obtained from non-empty axial slices. 

For our training pipeline we use the default Brain MRI configuration of the task from~\cite{MOOD22}.
We also adapted the same $\alpha_t$ schedule and network architecture proposed in DDPM but reduced the steps to $T=100$.
Our UNet has six downsampling blocks with  $[32,64,96,128,256,256]$ channels and attention layers from the 3rd block onwards.
Networks are trained using the AdamW optimizer and OneCycle learning rate with maximum rate of 1e-4 for 100,000 steps. 

\begin{figure}[h!]
    \centering
    \includegraphics[width=0.9\columnwidth]{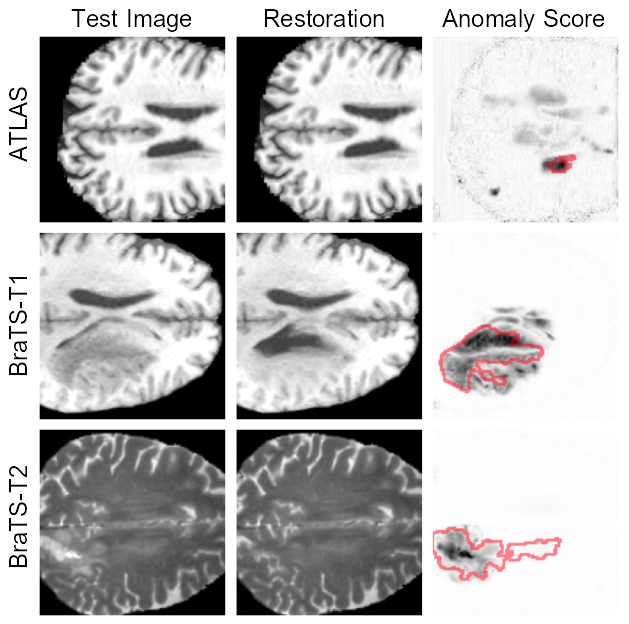}
    \caption{Restoration results on test images drawn from the ATLAS~\cite{ATLAS}, BraTS-T1, and BraTS-T2~\cite{brats}, after $step\_size=25$, along with anomaly scores $AS$ in grayscale compared with the ground truth outline in red.}
    \label{fig:qualitatives}
\end{figure}

\noindent\textbf{Performance analysis:} 
To quantitatively evaluate our method, we use the most common metrics in the medical UAD literature: Average Precision (AP) and an estimate of the best possible Sørensen-Dice index ([Dice] score). Some qualitative results are reported in Fig. \ref{fig:qualitatives}.

\begin{figure}[ht!]
    \centering
    \includegraphics[width=\columnwidth]{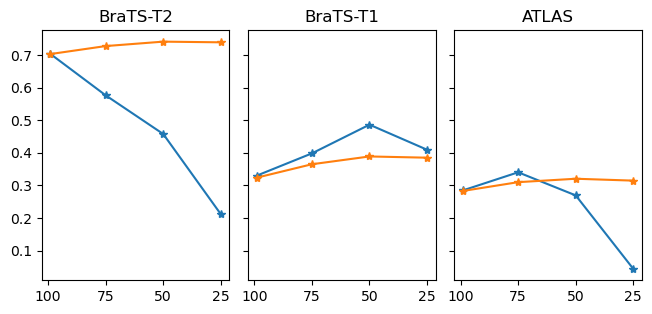}
    \caption{AP profiles as a function of $t$. Blue line shows single-step restoration. Orange line shows the cumulative AS as defined in Section \ref{sections:method} up-to $t$ stage ($step\_size = 25$). Single seed.}
    \label{fig:profiles}
\end{figure}

We evaluated single-step restorations to localize anomalies as a function of $t$, \emph{i.e.},  $\| x - P(x,t) \|$.
We also evaluated our cumulative multi-step restoration strategy using different $step\_size$ settings.
Results are included in Table~\ref{tab:ablation}. Additionally, Fig.~\ref{fig:profiles} compares single-step restoration AP with cumulative multi-step restoration performance.

\begin{table}[hbt]
    \centering
    \resizebox{0.90\columnwidth}{!}{
    \begin{tabular}{cccll}
   &$AP$ & \textbf{ATLAS}& \textbf{BraTS-T1}&\textbf{BraTS-T2}\\
        \midrule
         \multirow{4}{*}{\rotatebox[origin=c]{90}{\makecell{\textbf{single-step} \\ $t$}}} & 25 &0.03 ± 0.00&   0.38 ± 0.01 & 0.23 ± 0.01\\
         & 50 &0.20 ± 0.02&   \textbf{0.44} ± 0.02 & 0.45 ± 0.01\\
         & 75 &\textbf{0.29} ± 0.03&   0.34 ± 0.03 & 0.59 ± 0.03\\
         & 100 &0.25 ± 0.03&   0.28 ± 0.02&\textbf{0.70} ± 0.02\\
        \midrule
        \multirow{5}{*}{\rotatebox[origin=c]{90}{\makecell{\textbf{multi-step} \\ $step\_size$}}} 
         & 10& 0.27 ± 0.02& 0.32 ± 0.02& 0.73 ± 0.02\\
         & 20& 0.29 ± 0.02& 0.34 ± 0.02& 0.74 ± 0.02\\
         & 25&  0.29 ± 0.02&  0.34 ± 0.02&  0.75 ± 0.02\\
         & 33& 0.29 ± 0.02& 0.35 ± 0.02& 0.75 ± 0.01\\
         & 50&  0.30 ± 0.03&  0.35 ± 0.02&  0.75 ± 0.02\\
        \bottomrule
    \end{tabular}
    }
    \caption{Upper section: AP for single-step restorations ($AS = \| x - P(x,t) \|$) as a function of the $t$ used to condition the model. Lower section: AP for multi-step restorations, as a function of $step\_size$. Mean and std. dev. on 4 seeds.}
    \label{tab:ablation}
\end{table}

Finally, Table~\ref{tab:main} compares our results with baseline UAD methods from~\cite{UPD}. For our method DISYRE we report mean metrics for four training runs with multi-step restoration and $step\_size = 25$. In addition to our method we also evaluate the method from~\cite{MOOD22}, included in the table as \textit{MOOD22}. Note that this approach uses the same synthetic anomaly generation process but binary cross-entropy with $\alpha$ as soft target. When implementing this baseline we use the same network architecture without conditioning on $t$.  

\newcommand{\api}{AP\textsubscript{i}}
\newcommand{\app}{AP}
\newcommand{\dice}{$\lceil$Dice$\rceil$}
\newcommand{\row}[5]{}  

\begin{table}[hbt!]
    \centering
    \resizebox{\columnwidth}{!}{
    \begin{tabular}{clcccccc}
        & \multicolumn{1}{c}{} & \multicolumn{2}{c}{\textbf{ATLAS}} & \multicolumn{2}{c}{\textbf{BraTS-T1}} & \multicolumn{2}{c}{\textbf{BraTS-T2}} \\
        \cmidrule(lr){3-4} \cmidrule(lr){5-6} \cmidrule(lr){7-8}
        & Method & \app & \dice & \app & \dice & \app & \dice \\
        \midrule
        \multirow{3}{*}{\rotatebox[origin=c]{90}{\textbf{IR}}}
        & VAE \cite{Baur2020} & 0.11 & 0.20 & 0.13 & 0.19 & 0.28 & 0.33 \\
        & r-VAE \cite{Chen2020} & 0.09 & 0.17 & 0.13 & 0.19 & 0.36 & 0.40 \\
        & f-AnoGAN \cite{Schlegl2019} & 0.02 & 0.06 & 0.06 & 0.12 & 0.15 & 0.21 \\
        \midrule
        \multirow{2}{*}{\rotatebox[origin=c]{90}{\textbf{FM}}}
        & FAE \cite{FAE} & 0.08 & 0.18 & \textbf{0.42} & \textbf{0.45} & 0.51 & 0.52 \\
        & RD \cite{deng2022anomaly} & 0.11 & 0.22 & 0.36 & 0.42 & 0.47 & 0.50 \\
        \midrule
        \multirow{1}{*}{\rotatebox[origin=c]{90}{\textbf{AB}}}
        & AMCons \cite{consUPS} & 0.01 & 0.03 & 0.05 & 0.12 & 0.35 & 0.40 \\
        \midrule
        \multirow{4}{*}{\rotatebox[origin=c]{90}{\textbf{S-S}}}
        & PII \cite{PII} & 0.03 & 0.07 & 0.13 & 0.22 & 0.13 & 0.22 \\
        & DAE \cite{DAE} & 0.05 & 0.13 & 0.13 & 0.20 & 0.47 & 0.49 \\
        & MOOD22 \cite{MOOD22} & 0.10 & 0.21 & 0.24 & 0.31 & 0.47 & 0.48 \\
        & DISYRE (Ours) & 0.19 & 0.27 & 0.22 & 0.30 & 0.69 & 0.67 \\
        & DISYRE f.w. (Ours) & \textbf{0.29} & \textbf{0.37} & 0.34 & 0.41 & \textbf{0.75} & \textbf{0.70} \\
        \bottomrule
    \end{tabular}}
    \caption{Localization results of the image-reconstruction (\textbf{IR}), feature-modeling (\textbf{FM}), attention-based (\textbf{AB}), and self-supervised (\textbf{S-S}) methods. Best scores are bold.}
    \label{tab:main}
\end{table}

\section{Discussion \& Conclusion} 
\label{sections:discussion}

Results in Table \ref{tab:main} highlight the effectiveness of our strategy: DISYRE achieves an AP of 0.75 in BraTS-T2 dataset, an increase of 0.24 from the second best method. DISYRE also improves the state of the art of the ATLAS task to 0.29, up from 0.19 while achieving competitive results in BraTS-T1. 

When using single-step restorations we found that AP peaks at different $t$ values for different anomalies and modalities (Fig. \ref{fig:profiles}).
We hypothesize that these profiles are associated with how the anomalies manifest in each image modality, with high-frequency anomalies being restored at lower $t$ values and more structural corruptions restored as $t$ approaches $T$. 
We aim to be sensitive to all classes of anomaly, from structural to high frequency, which is why we integrate multiple steps into the proposed $AS$.
Despite improving results in BraTS-T2 and ATLAS, multi-step $AS$ does not improve single-step $AS$ in BraTS-T1, potentially due to the nature of how gliomas manifest in T1-weighted Brain MRI.
In future work we will explore alternative approaches to achieve better coverage of the anomaly spectrum.

As seen in Table \ref{tab:ablation}, DISYRE is robust to the choice of the $step\_size$ hyper-parameter.
We opted for $step\_size = 25$ in our final experiments as it provides a good balance between coverage of the anomaly schedule and inference speed.

In this work we propose DISYRE, a new strategy that adapts diffusion models to UAD.
With the motivation of learning a score function $\nabla_x \log p(x)$ which generalizes to naturally occurring medical anomalies, we replace the standard Gaussian noise corruption with gradual, synthetic anomalies.
DISYRE opens a new route to leverage diffusion-like models in medical image analysis; we show that forward processes can be designed so they become relevant for downstream tasks. In DISYRE we leverage synthetic anomalies for UAD and substantially outperform other UAD baseline methods in two out of three datasets.

\section{Compliance with Ethical Standards}
This research study was conducted retrospectively using human subject data made available in open access by \cite{CamCAN,ATLAS,brats}.
Ethical approval was not required as confirmed by the license attached with the open-access data.

\section{Acknowledgments}
\label{sec:acknowledgments}

Bernhard Kainz received support by the ERC - project MIA-NORMAL 101083647, Matthew Baugh by a UKRI DTP award.







\bibliographystyle{IEEEbib}
\bibliography{YADUMAS}

\begin{thebibliography}{10}

\bibitem{Litjens2017}
Geert Litjens et~al.,
\newblock ``A {Survey} on {Deep} {Learning} in {Medical} {Image} {Analysis},''
\newblock {\em Medical image analysis}, vol. vol. 42, pp. 60--88, 2017.

\bibitem{Baur2020}
Christoph Baur et~al.,
\newblock ``Autoencoders for {Unsupervised} {Anomaly} {Segmentation} in {Brain} {MR} {Images}: {A} {Comparative} {Study},''
\newblock {\em Medical image analysis}, , no. 02 Jan 2021, pp. 69:101952, 2021.

\bibitem{Schlegl2019}
Thomas Schlegl et~al.,
\newblock ``f-{AnoGAN}: {Fast} unsupervised anomaly detection with generative adversarial networks,''
\newblock {\em Medical image analysis}, vol. 54, pp. 30--44, 2019.

\bibitem{phanes}
Cosmin~I. Bercea et~al.,
\newblock ``Reversing the {Abnormal}: {Pseudo}-{Healthy} {Generative} {Networks} for {Anomaly} {Detection},''
\newblock in {\em Medical {Image} {Computing} and {Computer} {Assisted} {Intervention} – {MICCAI} 2023}, Cham, 2023, pp. 293--303.

\bibitem{Chen2020}
Xiaoran Chen et~al.,
\newblock ``Unsupervised lesion detection via image restoration with a normative prior,''
\newblock {\em Medical Image Analysis}, vol. 64, pp. 101713, Aug. 2020.

\bibitem{Naval2020}
Sergio Naval~Marimont and Giacomo Tarroni,
\newblock ``Anomaly detection through latent space restoration using vector quantized variational autoencoders,''
\newblock in {\em 2021 {IEEE} 18th {International} {Symposium} on {Biomedical} {Imaging} ({ISBI})}, 2021.

\bibitem{DDPM}
Jonathan Ho et~al.,
\newblock ``Denoising diffusion probabilistic models,''
\newblock {\em Advances in neural information processing systems}, vol. 33, pp. 6840--6851, 2020.

\bibitem{anoddpm}
Julian Wyatt et~al.,
\newblock ``{AnoDDPM}: {Anomaly} {Detection} with {Denoising} {Diffusion} {Probabilistic} {Models} using {Simplex} {Noise},''
\newblock in {\em 2022 {IEEE}/{CVF} {Conference} on {Computer} {Vision} and {Pattern} {Recognition} {Workshops} ({CVPRW})}, New Orleans, LA, USA, June 2022, pp. 649--655.

\bibitem{DAE}
Antanas Kascenas et~al.,
\newblock ``Denoising autoencoders for unsupervised anomaly detection in brain mri,''
\newblock in {\em International Conference on Medical Imaging with Deep Learning}. PMLR, 2022, pp. 653--664.

\bibitem{WollebJ20222}
Julia Wolleb et~al.,
\newblock ``Diffusion {Models} for {Medical} {Anomaly} {Detection},''
\newblock {\em arXiv preprint arXiv:2203.04306}.

\bibitem{fpi}
Jeremy Tan et~al.,
\newblock ``Detecting outliers with foreign patch interpolation,''
\newblock {\em Machine Learning for Biomedical Imaging}, vol. 1, no. April 2022 issue, pp. 1--27, 2022.

\bibitem{PII}
Jeremy Tan et~al.,
\newblock ``Detecting {Outliers} with {Poisson} {Image} {Interpolation},''
\newblock in {\em Medical {Image} {Computing} and {Computer} {Assisted} {Intervention} – {MICCAI} 2021}, Cham, 2021.

\bibitem{MOOD22}
Sergio Naval~Marimont and Giacomo Tarroni,
\newblock ``Achieving state-of-the-art performance in the {Medical} {Out}-of-{Distribution} ({MOOD}) challenge using plausible synthetic anomalies,'' Nov. 2023,
\newblock arXiv:2308.01412 [cs].

\bibitem{cold_diffusion}
Arpit Bansal et~al.,
\newblock ``Cold {Diffusion}: {Inverting} {Arbitrary} {Image} {Transforms} {Without} {Noise},'' Aug. 2022,
\newblock arXiv:2208.09392 [cs].

\bibitem{UPD}
Ioannis Lagogiannis et~al.,
\newblock ``Unsupervised {Pathology} {Detection}: {A} {Deep} {Dive} {Into} the {State} of the {Art},''
\newblock {\em IEEE Transactions on Medical Imaging}, pp. 1--1, 2023,
\newblock arXiv:2303.00609 [cs].

\bibitem{CamCAN}
Jason~R Taylor et~al.,
\newblock ``The {Cambridge} {Centre} for {Ageing} and {Neuroscience} ({Cam}-{CAN}) data repository: {Structural} and functional {MRI}, {MEG}, and cognitive data from a cross-sectional adult lifespan sample,''
\newblock {\em neuroimage}, vol. 144, pp. 262--269, 2017.

\bibitem{ATLAS}
Sook-Lei Liew et~al.,
\newblock ``A large, open source dataset of stroke anatomical brain images and manual lesion segmentations,''
\newblock {\em Scientific data}, vol. 5, pp. 1--11, 2018.

\bibitem{brats}
Bjoern~H Menze et~al.,
\newblock ``The multimodal brain tumor image segmentation benchmark (brats),''
\newblock {\em IEEE transactions on medical imaging}, vol. 34, no. 10, pp. 1993--2024, 2014.

\bibitem{FAE}
Felix Meissen et~al.,
\newblock ``Unsupervised {Anomaly} {Localization} with {Structural} {Feature}-{Autoencoders},''
\newblock in {\em Brainlesion: {Glioma}, {Multiple} {Sclerosis}, {Stroke} and {Traumatic} {Brain} {Injuries}}, Cham, 2023, pp. 14--24.

\bibitem{deng2022anomaly}
Hanqiu Deng and Xingyu Li,
\newblock ``Anomaly detection via reverse distillation from one-class embedding,''
\newblock in {\em Proceedings of the IEEE/CVF Conference on Computer Vision and Pattern Recognition}, 2022, pp. 9737--9746.

\bibitem{consUPS}
Julio Silva-Rodríguez et~al.,
\newblock ``Constrained unsupervised anomaly segmentation,''
\newblock {\em Medical Image Analysis}, vol. 80, pp. 102526, Aug. 2022,
\newblock arXiv:2203.01671 [cs, eess].

\end{thebibliography}

\end{document}